\documentclass[times, twoside]{henriques-style}
\usepackage{blindtext}
\usepackage[nolist]{acronym}
\setcitestyle{square}
\usepackage{tikz}
\usetikzlibrary{positioning, calc, matrix, decorations.pathreplacing, shapes.geometric}
\usepackage{adjustbox}

\usepackage{xspace}
\makeatletter
\DeclareRobustCommand\onedot{\futurelet\@let@token\@onedot}
\def\@onedot{\ifx\@let@token.\else.\null\fi\xspace}

\leadauthor{Halmes} 

\begin{document}

\title{Deep Learning-based mitosis detection in breast cancer histologic samples}
\shorttitle{Deep Learning-based mitosis detection in breast cancer histologic samples}

% Use letters for affiliations, numbers to show equal authorship (if applicable) and to indicate the corresponding author
\author{Michel Halmes}
\author{Hippolyte Heuberger}
\author{Sylvain Berlemont}

\affil[1]{Keen Eye Technologies S.A.S.}

\maketitle

%TC:break Abstract
%the command above serves to have a word count for the abstract
\begin{abstract}
This is the submission for mitosis detection in the context of the MIDOG 2021 challenge~\cite{midog}. It is based on the two-stage objection model Faster RCNN~\cite{fasterrcnn} as well as DenseNet~\cite{densenet} as a backbone for the neural network architecture. It achieves a F1-score of 0.6645 on the Preliminary Test Phase Leaderboard.    
\end {abstract}
%TC:break main
%the command above serves to have a word count for the abstract

%\begin{keywords}
%Domain Generalization | Mitotic Count | Histopathology
%\end{keywords}

\begin{corrauthor}
hippolyte.heuberger@keeneye.ai
\end{corrauthor}

\section*{Introduction}
Prognosis of breast cancer is assessed using multiple methods, among which is the estimation of the tumor proliferation, corresponding to how fast the tumor is growing. The tumor proliferation can be estimated from the H\&E sample by quantifying the number of mitotic figures in the sample. In order to limit intra and inter observer variability in the quantification, an automatic method for detection of the mitotic figures is necessary. 
We build an object detection system for mitotic figures on H\&E-stained tissue of breast cancer. We use Faster-RCNN as the neural network architecture of the detector. In this report, we describe our approach and the obtained results.

\section*{Method}

\subsection*{Dataset preparation}
The dataset contains 150 slides. We randomly split out the dataset into 3 parts: 105 slides go into the training-set. 15 slides go into the validation-set which is used to optimize  hyperparameters. The remaining 30 are in the testing-set which we use to report our evaluation metrics on. In a second phase we study the impact of training on two scanner and evaluating on the third one. These two data preparation methods are only used to design the model, in the end all data available is used as a training set.

Our slides have heights and widths between 5000 and 7000 pixels. Such images are of course too large to feed into a neural network directly.

Therefore, we divide all slides into tiles of size 1024. Most slides contain between 35 and 48 tiles. 
The boxes around our cells undergoing mitosis have a standardized size of 50x50. For this reason, we work with the images at the original scale-factor (i.e. without scaling them down).

The dataset also contains annotations for non-mitotic figures. Those annotations are a cue not interesting for our challenge. We still decided to keep these annotations at training time to detect this second class of objects. This provides additional information at training time and teaches the detector to explicitly make the distinction with cells that are not undergoing mitosis.

In our total of 6263 tiles, there are only 1171 cells undergoing mitosis. With this data distribution, the model will naturally converge to the trivial solution of classifying everything as background. Therefore, at training time, we drop tiles that do not contain any object with 80\% probability.

\subsection*{Data augmentation}
The annotations we received are on images from 3 different scanners. The final evaluation will however, contain images from 6 different scanners. 
Each scanner has its own colorimetric properties. This means that for the same piece of tissue will result in different images. They will have a different hue for the hematoxylin and the eosin, different saturation and brightness.

The challenge here is to prepare the detector to be able to perform well on unseen scanners. For this we use data augmentation, mostly those that reproduce the variability between scanners.

Below, we describe the augmentations used. We do not use all augmentations at the same time. For each of them we specify a probability of augmentation to be activated. This makes sure that it is still possible for the image to be completely unchanged.
\begin{itemize}
\item \textbf{Brightness augmentation}:  Brightness is adjusted by adding a delta to the RGB values. The delta is uniformly sampled between -0.2 and +0.2. We activate this augmentation with 20\% probability.
\item \textbf{Hue augmentation}: We add an offset delta to the hue channel. The delta is uniformly sampled between -0.1 and +0.1. Again, it is activated with 20\% probability.
\item \textbf{Contrast augmentation}: We adjust the contrast of our images by a factor sampled between -20\% and + 20\%. Again, it is activated with 20\% probability.
\item \textbf{Saturation augmentation}: We multiply the saturation channel of our images by a factor sampled between -20\% and + 20\%. Again, it is activated with 20\% probability.
\item \textbf{Symmetric flip}: We add another non-colorimetric augmentation to enrich the data. This augmentation flips the image around the horizontal and vertical central axis. We select randomly whether to flip around each axis with equal probability. This also means that there is a 25\% chance of the image not being modified.
\end{itemize}

\subsection*{Model}
We use Faster RCNN~\cite{fasterrcnn} as object detector combined with DenseNet-121~\cite{densenet} as backbone.
We make 3 adjustment to the standard parameters:

All anchor scales are set to 50 pixels, since all boxes that have to be detected are at that size. This means that we detect the same scale of objects at various scales of feature maps.
We only look for objects with anchor ratio 1: all objects we look for are square. This improvement was done as a speed improvement, rather than for performance itself.
Mitotic figures are very rare objects, as we already mentioned above. We need to teach the model that most tissue is just background. For this reason, we decrease the fraction of positive anchors selected from 50\% in the default implementation to 25\%.

\subsection*{Post-processing}
There are two important steps after the output of the model.

Non-maximum suppression removes detections with a too high overlap. Since the evaluation metric uses an IOU of 10\%, we use this threshold here as well.

Secondly, we select a threshold for the model's confidence score, below which we consider the detections to contain background.
The threshold was determined automatically by grid-search on the whole set of images to maximize the F1-score.

\section*{Results}
For the choice of the methodology, the neural network structure as well as hyper parameter tuning, the evaluation set was a 3-fold cross validation where two scanners were used for training while the third was kept unseen and only used for evaluation. The final model is trained on the whole dataset made available and achieves on the Preliminary Test Phase Leaderboard the \textbf{F1-score of 0.6645}. The associated recall is 0.6084 and the precision is 0.7319.

\section*{Discussion and Conclusion}
The presented approach ends up sharing similarities to the approach used for annotating the data, based on training using all labels made at disposal, including false positives. A large focus in this study has been however put on the data augmentation, to ensure that images coming from a given scanner present characteristics as similar as possible from images from the other scanners.
While multiple different additional modules were tested (such as a different object detection architecture, different backbones, or style transfer) to make the methodology more comprehensive, the quantitative experiments we ran suggested keeping the methodology as simple as possible.
On the private leaderboard this results in a F1 score of 0.6645 which remains modest from an object detection point of view. Nevertheless the use of such a tool would be beneficial to guide / complete manual assessment and would hopefully contribute to standardise the estimation of tumor proliferation.

\section*{Bibliography}
\bibliography{bibliography}

\begin{acronym}
\acro{midog}[MIDOG]{MItosis DOmain Generalization}
\acro{he}[H\&E]{Hematoxylin \& Eosin}
\end{acronym}

\end{document}